%% file: arxiv.tex
\pdfoutput=1
\documentclass[10pt,twocolumn,letterpaper]{article}

\usepackage{text}             
\input{preamble}
\usepackage{xcolor}
\usepackage{soul}

\definecolor{Palette1}{RGB}{78,101,155}   
\definecolor{Palette2}{RGB}{138,140,191}  
\definecolor{Palette3}{RGB}{184,168,207}  
\definecolor{Palette4}{RGB}{231,188,198}  
\definecolor{Palette5}{RGB}{253,207,158}  
\definecolor{Palette6}{RGB}{239,164,132}  
\definecolor{Palette7}{RGB}{182,118,108}  
\definecolor{cvprblue}{rgb}{0.21,0.49,0.74}
\definecolor{SPARK_yellow}{rgb}{253,207,158}

\usepackage[pagebackref,breaklinks,colorlinks,allcolors=cvprblue]{hyperref}

\newcommand\blfootnote[1]{%
  \begingroup
  \renewcommand\thefootnote{}\footnote{#1}%
  \addtocounter{footnote}{-1}%
  \endgroup
}

\title{SPARK: Sim-ready Part-level Articulated Reconstruction with VLM Knowledge}

\author{
Yumeng He$^{1,2*}$  \quad
Ying Jiang$^{1*}$ \quad 
Jiayin Lu$^{1*}$ \quad 
Yin Yang$^{3}$ \quad
Chenfanfu Jiang$^{1}$
}

\begin{document}

\setcounter{footnote}{0}

\twocolumn[{%
\renewcommand\twocolumn[1][]{#1}%
\maketitle

\begin{center}
    \centering
    \captionsetup{type=figure}
    \includegraphics[width=\textwidth, trim=0 0 0 0, clip]{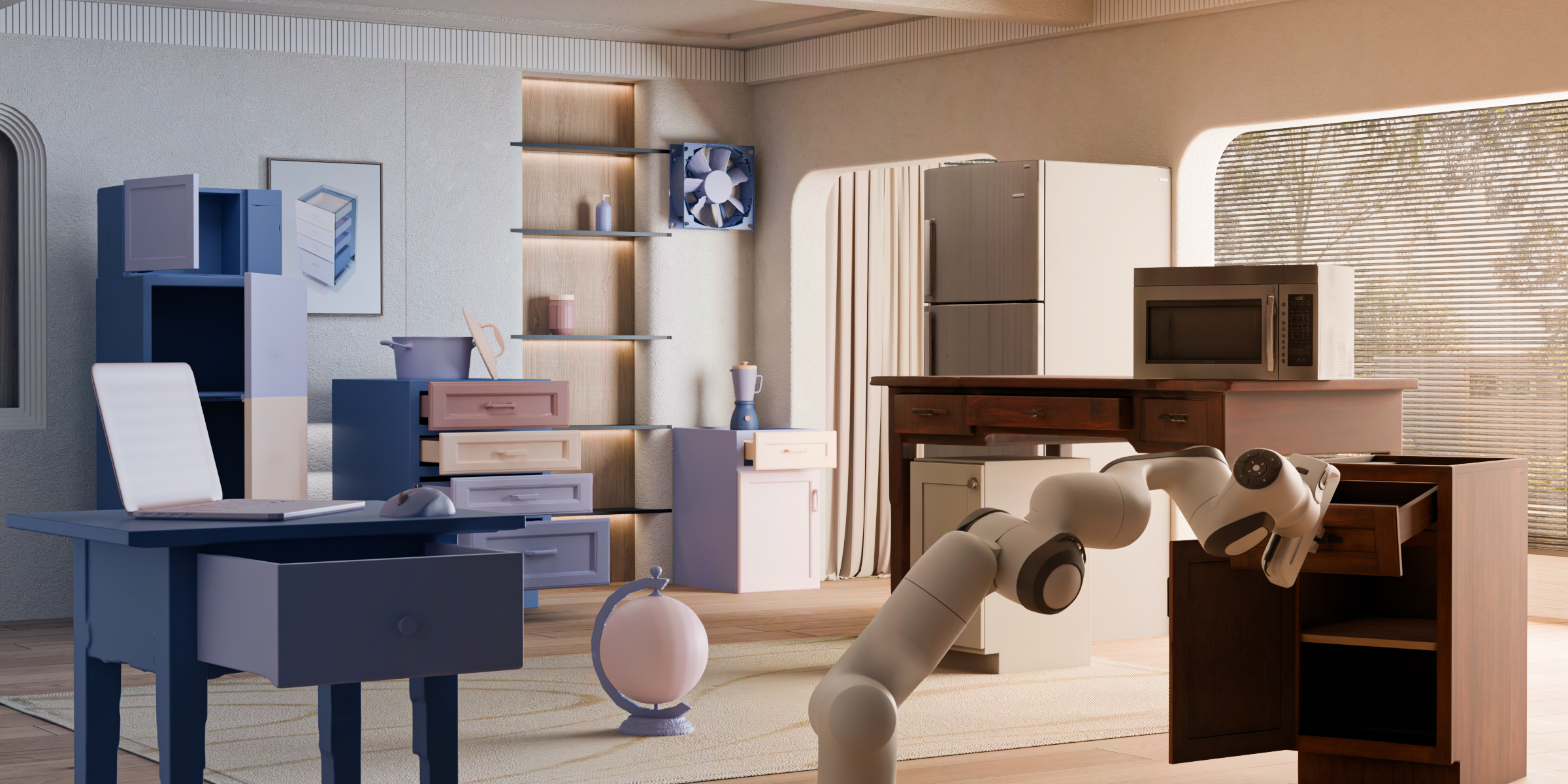}
    \captionof{figure}{\textbf{SPARK} is a novel framework that integrates VLM-guided part-level and global image guidance with diffusion transformers to produce high-quality articulated object reconstructions.}
    \label{fig:teaser}
\end{center}
}]
\blfootnote{* equal contribution. $^{1}$ UCLA, 
$^{2}$ USC,
$^{3}$ University of Utah. \url{heyumeng@usc.edu}, 
\url{jiayin\_lu, yingjiang, cffjiang@ucla.edu}, \url{yin.yang@utah.edu}
}
\input{sec/0_abstract}    
\input{sec/1_intro}

\input{sec/2_related}
\input{sec/3_method}

\input{sec/4_experiments}

\input{sec/5_conclusion}

{
    \small
    \bibliographystyle{ieeenat_fullname}
    \bibliography{main}
}

\newpage
\appendix
\section*{\Large Appendix}

\input{sec/arxiv_sup}

\end{document}

%% file: preamble.tex
\usepackage[table]{xcolor}
\usepackage{wrapfig}

%% file: sec/0_abstract.tex
\begin{abstract}
Articulated 3D objects are critical for embodied AI, robotics, and interactive scene understanding, yet creating simulation-ready assets remains labor-intensive and requires expert modeling of part hierarchies and motion structures. We introduce SPARK, a framework for reconstructing physically consistent, kinematic part-level articulated objects from a single RGB image. Given an input image, we first leverage VLMs to extract coarse URDF parameters and generate part-level reference images. We then integrate the part-image guidance and the inferred structure graph into a generative diffusion transformer to synthesize consistent part and complete shapes of articulated objects. To further refine the URDF parameters, we incorporate differentiable forward kinematics and differentiable rendering to optimize joint types, axes, and origins under VLM-generated open-state supervision. Extensive experiments show that SPARK produces high-quality, simulation-ready articulated assets across diverse categories, enabling downstream applications such as robotic manipulation and interaction modeling. Project page: \url{https://heyumeng.com/SPARK/index.html}.

\end{abstract}

%% file: sec/1_intro.tex
\vspace{-15px}
\section{Introduction}
\label{sec:intro}
\vspace{-5px}

Articulated objects are ubiquitous in everyday environments, and scalable creation of high-fidelity, interactable 3D assets is becoming increasingly important for embodied AI, robotics, and scene reconstruction. However, building articulated 3D models remains labor-intensive, often requiring manual labeling and expert modeling of part objects. With the advance of powerful generative models \cite{liu2023zero, lin2023magic3d}, it is now possible to synthesize high-quality 3D assets directly from images or text prompts. However, the generated shapes are typically fused, making them difficult to reuse for downstream manipulation, animation, and simulation. Recent progress in part-level generation \cite{lin2025partcrafter, chen2025autopartgen, tang2024partpacker} enables the synthesis of semantically meaningful 3D parts, supporting fine-grained editing and control. Yet, despite strong geometric quality, these models often lack kinematic consistency because their part segmentation is driven purely by appearance, ignoring the underlying motion structure.

Recently, articulated object generation methods~\cite{liu2025artgs, chen2025freeart3d, qiu2025articulate, le2024articulate, gao2025partrm, gao2025meshart, su2025artformercontrollablegenerationdiverse} have sought to synthesize kinematic part-level objects together with their kinematic parameters, including joint and link properties. Many approaches~\cite{liu2025artgs, chen2025freeart3d, qiu2025articulate, le2024articulate, gao2025partrm} first generate or retrieve a 3D mesh and then segment it into movable parts under motion or multi-state image guidance, but these methods depend on template meshes or require multiple images for reliable segmentation. Other works~\cite{gao2025meshart, guo2025kinematic, su2025artformercontrollablegenerationdiverse} directly generate articulated objects using kinematic graphs as structural priors, while they require explicit kinematic parameters as input. For full URDF synthesis, both optimization-based~\cite{lu2025dreamartgeneratinginteractablearticulated, liu2025artgs, gao2025partrm} and feed-forward~\cite{li2025urdf, le2024articulate, chen2024urdformer, wu2025dipodualstateimagescontrolled} methods have been proposed to estimate articulation parameters from multi-state images, videos, meshes, or reconstructed point clouds. To minimize user input, our goal is to reconstruct both kinematic part-level articulated objects and their corresponding URDF parameters from a single image.

To address these challenges, we present SPARK, a framework that reconstructs simulation-ready articulated objects at the kinematic part level and estimates complete URDF parameters from a single RGB image. Given an input image, we first leverage vision–language models (VLMs) to extract coarse URDF parameters for joints and links. We then generate part-level images and construct a structure graph that encodes parent–child link relationships. Conditioning on both the structure graph and the synthesized part images, SPARK produces consistent part-level geometry and assembles them into a coherent articulated mesh through multi-level hierarchical attention. Finally, to further refine URDF parameters, we incorporate differentiable forward kinematics and differentiable rendering with a feature-injection strategy, optimizing joint attributes under the supervision of VLM-generated open-state images. To summarize, our contributions are:

\begin{itemize}
    \item We propose a novel framework that integrates a generative diffusion transformer with VLM priors to synthesize high-quality kinematic part-level articulated objects and accurately estimate the corresponding URDF parameters.

    \item We introduce part-image guidance and multi-level attention to enable consistent multi-part synthesis, along with a joint optimization component that refines kinematic parameters under VLM-guided supervision.

    \item Extensive experiments demonstrate that SPARK produces accurate and simulation-consistent articulated assets across diverse object categories.
\end{itemize}

%% file: sec/2_related.tex
\section{Related Work}
\label{sec:related}

\vspace{-3px}
\subsection{3D Object Generation}
\vspace{-3px}
\label{sec:3Dgen}
3D object generation aims to synthesize 3D shapes from texts, images, and point clouds, producing various representations such as triangle meshes \cite{xiang2024structured, hunyuan3d2025hunyuan3d, chen2025ultra3defficienthighfidelity3d, wu2024unique3d, li2025triposg, zhao2024di-pcg}, analytic primitives \cite{ye2025primitiveanything, ma2024parameterizestructuredifferentiabletemplate}, point clouds \cite{romanelis2024efficientscalablepointcloud, lan2024ga}, signed distance fields (SDFs) \cite{antic2025sdfit}, neural radiance fields (NeRFs) \cite{Erkoc_2023_ICCV, wang2023prolificdreamer, 10422989}, and 3D Gaussian splats (3DGS) \cite{Ren_2024_CVPR, tang2023dreamgaussian, roessle2024l3dg, chen2024textto3dusinggaussiansplatting, yi2023gaussiandreamer}. Among these, mesh generation is especially appealing for its high fidelity and compatibility with simulation and graphics pipelines \cite{li2025triposg, xiang2024structured, zhao2025hunyuan3d}. To this end, TripoSG \cite{li2025triposg} and TRELLIS \cite{xiang2024structured} employ flow-based models to generate high-fidelity 3D shapes, while Hunyuan3D \cite{hunyuan3d2025hunyuan3d} adopts a Diffusion Transformer (DiT) that first generates geometry and then recovers textures. However, the generated shapes are typically fused, requiring additional 3D segmentation. SAMPart3D \cite{yang2024sampart3d} and PartDistill \cite{umam2023partdistill} adopt 2D-to-3D distillation strategies to segment 3D shapes from 2D priors, though the results are often incomplete and geometrically inconsistent in occluded regions. To further obtain complete 3D parts, HoloPart \cite{yang2025holopart} employs a dual-attention diffusion model to complete partial segments, while PartField \cite{partfield2025} leverages labeled 3D supervision to learn part features extending into the object interior. Instead of generating a complete shape followed by part extraction, recent approaches directly synthesize 3D parts from image latent representations \cite{tang2024partpacker, chen2025autopartgen, lin2025partcrafter, yang2025omnipart, liu2025partcomposer}. PartComposer \cite{liu2025partcomposer} and OmniPart \cite{yang2025omnipart} use spatial bounding boxes and 2D segmented part images, respectively, to guide part-level object generation with latent flow models and diffusion models. To further enhance structural guidance, AutoPartGen \cite{chen2025autopartgen} exploits an autoregressive latent flow transformer to generate parts sequentially. In contrast, PartCrafter~\cite{lin2025partcrafter} and DualPacker~\cite{tang2024partpacker} generate all parts simultaneously using a DiT with part-wise latents or a rectified-flow model with dual latent volumes. However, the resulting parts often lack kinematic awareness, leading to over- or under-segmentation. Our goal is to generate articulated objects at the kinematic-part level from a single image.

\vspace{-3px}
\subsection{Articulated Object Shape Reconstruction}
\vspace{-3px}
Articulated object shape reconstruction focuses on recovering kinematic part-level geometry for each movable component. Prior works typically first obtain a complete 3D representation either by generating a whole 3D model \cite{lu2025dreamartgeneratinginteractablearticulated, chen2025freeart3d, gao2025partrm, liu2025artgs, vora2025articulate} or by retrieving an existing 3D mesh \cite{qiu2025articulate} and then extract part-level articulated structures. Among these methods, Articulate AnyMesh \cite{qiu2025articulate} and DreamArt \cite{lu2025dreamartgeneratinginteractablearticulated} explicitly segment the 3D representation using segmentation models \cite{yang2025holopart, kirillov2023segment}, making their results sensitive to segmentation quality. While FreeArt3D \cite{chen2025freeart3d}, ArtGS \cite{liu2025artgs}, PartRM \cite{gao2025partrm}, and ATOP \cite{vora2025articulate} perform implicit part decomposition guided by SDS-based optimization, articulable embeddings, multi-state Gaussian fields, and motion embeddings, respectively. However, these methods may produce incomplete parts in occluded regions when decomposing whole 3D shapes.
On the other hand, Kinematic Kitbashing \cite{guo2025kinematic} retrieves articulated parts and optimizes their assembly according to structure graphs. In contrast, some methods generate part-level articulated objects and complete meshes simultaneously from structural parameters \cite{gao2025meshart, su2025artformercontrollablegenerationdiverse, liu2024cage} or from multi-view images \cite{liu2023paris, shen2025gaussianart}, using feed-forward generative networks or differentiable optimization frameworks. In contrast, our method focuses on simultaneously generating complete kinematic part-level articulated objects and the composed mesh from an image by leveraging VLM priors within an end-to-end framework.

\vspace{-3px}
\subsection{Articulation Parameter Estimation}
\vspace{-3px}
Articulation parameter estimation aims to recover an articulated object's kinematic graph, including both link and joint information \cite{featherstone2008rigid}, representable in the Unified Robot Description Format (URDF) \cite{quigley2015programming}, Simulation Description Format (SDF) \cite{sdformat}, etc., for downstream simulation and manipulation \cite{Todorov2012MuJoCoAP, Xiang_2020_SAPIEN, coleman2014reducing}.
To estimate articulation parameters, recent methods~\cite{li2025urdf, mandi2024real2codereconstructarticulatedobjects, chen2024urdformer, wu2025dipodualstateimagescontrolled} train end-to-end models on URDF-annotated data to directly predict URDF parameters from multi-state visual observations.
Specifically, Articulate-Anything \cite{le2024articulate} and Real2Code \cite{mandi2024real2codereconstructarticulatedobjects} leverage multimodal large language models (MLLMs) or large language models (LLMs) to infer URDF graphs or parameters from point clouds, images, or oriented bounding boxes, while URDFormer \cite{chen2024urdformer} and DIPO \cite{wu2025dipodualstateimagescontrolled} predict URDF structures from images using Transformer-based architectures.
In contrast, optimization-based methods such as DreamArt~\cite{lu2025dreamartgeneratinginteractablearticulated} and ArticulateGS~\cite{guo2025articulatedgs} explicitly optimize URDF joint parameters under multi-state video supervision, whereas Part$^2$GS~\cite{yu2025part2gspartawaremodelingarticulated}, GAMMA \cite{yu2024gammageneralizablearticulationmodeling}, and ArtGS~\cite{liu2025artgs} optimize motion fields from open–close image pairs, from which the corresponding URDF parameters are subsequently recovered. Unlike methods that optimize from scratch, Articulate-Anything~\cite{le2024articulate} and DreamArt~\cite{lu2025dreamartgeneratinginteractablearticulated} first generate URDF-like code and then refine it using VLM-based semantic feedback and generative video supervision. Inspired by this strategy, we use a VLM-generated URDF template and refine joint parameters via differentiable forward kinematics and rendering, guided by synthesized open–close image pairs.

%% file: sec/3_method.tex
\section{Method}
\label{sec:method}
\vspace{-5px}
\begin{figure*}[t]
    \centering
        \includegraphics[width=\textwidth]{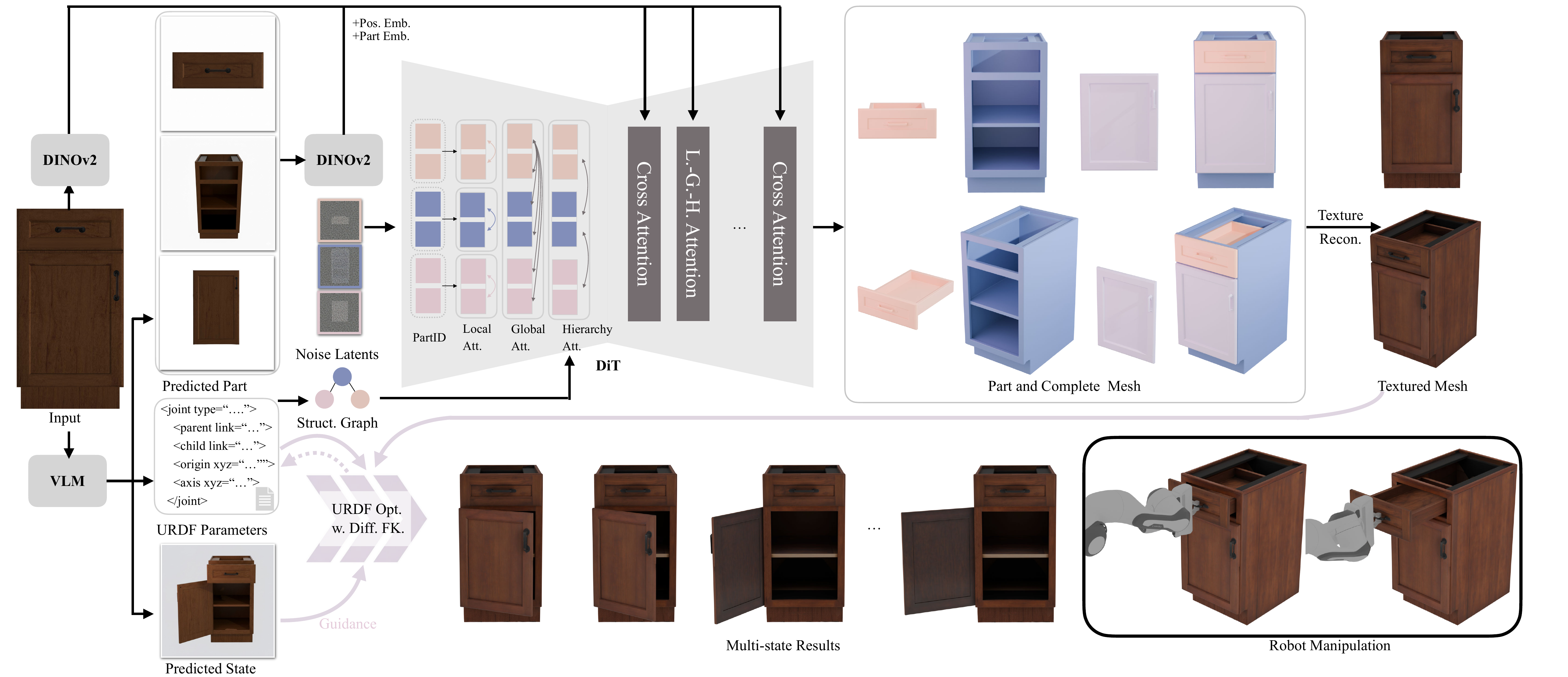}
        \vspace{-0.3in}
    \caption{\small{\textbf{Pipeline Overview.}
We use a VLM to generate per-part reference images, predicted open-state images, and URDF templates with preliminary joint and link estimations. A Diffusion Transformer (DiT) equipped with local, global, and hierarchical attention mechanisms simultaneously synthesizes part-level and complete articulated meshes from a single image with VLM priors. We further employ a generative texture model to generate realistic textures and refine the URDF parameters using differentiable forward kinematics and differentiable rendering under the guidance of the predicted open-state images.
}}\label{fig:pipeline}
\end{figure*}

Given an input image $I_0$, our goal is to generate an articulated object represented by a sequence of part-level meshes $\{\mathbf{M}_k\}_{k=1}^{K}$ that together form a composite whole mesh $\mathbf{M}$, and to estimate its hierarchical URDF parameters $\mathbf{u} = \{\mathbf{u}_\ell, \mathbf{u}_j\}$. 
Here, $\mathbf{u}_\ell$ represents the link nodes composed of the part meshes, while $\mathbf{u}_j$ encodes the kinematic information, including the joint type ${\mathbf{u}_{j}^{\text{type}}}$, joint axis ${\mathbf{u}_{j}^{\text{axis}}}$, joint origin ${\mathbf{u}_{j}^{\text{origin}}}$, and joint limits ${\mathbf{u}_{j}^{\text{limit}}}$. 
To achieve this, we introduce \textbf{SPARK}, a novel framework that integrates DiT with VLM priors and a differentiable optimization module. 
Starting from the input image $I_0$, the VLM-guided parsing stage first produces coarse URDF parameters $\mathbf{u}$, a set of per-part reference images $\{\mathbf{r}_k\}_{k=1}^{K}$, and a predicted open-state image $I_{\text{open}}$. 
These part images, together with the global input image and the inferred semantic link hierarchy, are integrated with a DiT to predict the 3D part-level meshes $\{\mathbf{M}_k\}_{k=1}^{K}$ and the complete articulated object $\mathbf{M}$, followed by texture generation. Based on the generated articulated objects, we further refine the joint parameters $\mathbf{u}j$ using differentiable forward kinematics combined with differentiable rendering and a feature-injection strategy, supervised by the predicted open-state image $I{\text{open}}$.

\vspace{-3px}
\subsection{VLM-guided Structural Reasoning}
\vspace{-3px}
In this section, we describe how to extract part images $\{\mathbf{r}_k\}_{k=1}^{K}$, and coarse URDF parameters $\mathbf{u}$, including link information $\mathbf{u}_\ell$ and joint attributes ${\mathbf{u}_{j}^{\text{type}}}$, ${\mathbf{u}_{j}^{\text{axis}}}$, ${\mathbf{u}_{j}^{\text{origin}}}$, and ${\mathbf{u}_{j}^{\text{limit}}}$, from a single input image $I_0$ through a VLM-guided parsing process. 

\vspace{-10px}
\paragraph{VLM-guided URDF Parameter Generation.}
The VLM first infers the part hierarchy, including the number and connectivity of links and joints, contributing to a URDF template consistent with the predicted structure. The resulting template instantiates the standard URDF schema by declaring links $\mathbf{u}_\ell$, parent--child relations, and joint specifications $\mathbf{u}_j = \{{\mathbf{u}_{j}^{\text{type}}}, {\mathbf{u}_{j}^{\text{axis}}}, {\mathbf{u}_{j}^{\text{origin}}}, {\mathbf{u}_{j}^{\text{limit}}}\}$. For link nodes $\mathbf{u}_\ell$, the VLM assigns semantic identifiers or names, which are later used to associate each link with its corresponding part mesh. For joint parameters $\mathbf{u}_j$, we categorize the attributes into discrete and continuous types. The discrete attributes are selected by the VLM from predefined dictionaries to ensure semantic and directional consistency. For instance, ${\mathbf{u}_{j}^{\text{type}}} \in \{\textit{fixed}, \textit{revolute}, \textit{prismatic}\}$ defines the joint category, whereas ${\mathbf{u}_{j}^{\text{axis}}}$ is selected from a canonical set of unit directions: front $(0,0,1)$, back $(0,0,-1)$, up $(0,1,0)$, down $(0,-1,0)$, right $(1,0,0)$, and left $(-1,0,0)$. For prismatic joints, the mapping directly corresponds to these translational directions. For instance, if a drawer is expected to be pulled forward, we assign $(0,0,1)$. For revolute joints, the mapping is determined by first identifying the rotation axis (among $x$, $y$, or $z$) and then determining its sign. We follow the standard convention in which clockwise rotation is negative and counterclockwise rotation is positive. For example, if a door opens toward the left, it rotates around the $y$-axis in a clockwise direction, which we define as $(0,-1,0)$. Another discrete attribute is the joint motion limit ${\mathbf{u}_{j}^{\text{limit}}}$, which consists of two values: \textit{lower} and \textit{upper}. The VLM estimates these limits, where the lower bound is always set to 0. For the upper bound, we use predefined values that differ between large-range motions (e.g., doors or drawers) and small-range motions (e.g., buttons). The continuous attributes, including the joint origin ${\mathbf{u}_{j}^{\text{origin}}}$ and joint motion limits ${\mathbf{u}_{j}^{\text{limit}}}$, are coarsely estimated by the VLM as initial positions and thresholds. 

\vspace{-10px}
\paragraph{Part Image and Structural Guidance.}
After obtaining the coarse URDF parameters, we extract the part count $K$ and the semantic category of each link (e.g., drawer, door, frame). Based on these semantic labels, we generate a set of per-part reference images $\{\mathbf{r}_k\}_{k=1}^{K}$, where each $\mathbf{r}_k$ corresponds to a single semantic part. We then construct a structural graph according to the parent–child relationships defined in the coarse URDF parameters. The generated part images and structural graph jointly serve as guidance for part-level generation in Sec.~\ref{sec:part_generation}.

\subsection{Part-Articulated Object Generation}
\vspace{-3px}
\label{sec:part_generation}
 Given per–part reference images $\mathcal{R}=\{\mathbf{r}_k\}_{k=1}^{K}$ and a single global image $I_0$, this stage reconstructs a set of $K$ geometry latents, one per articulated part, that decode into a part-decomposable, simulation-ready 3D asset. We adopt a Diffusion-Transformer (DiT), itself inspired by TripoSG, and condition denoising on both local (per-part) and global (whole-object) visual evidence. 

\vspace{-10px}
\paragraph{Multi-Level Attention Mechanisms.}

We begin by extracting visual guidance for each part. A shared DINOv2 encoder~\cite{oquab2023dinov2} maps every part image $\mathbf{r}_k$ and the replicated global image $I_0$ into token sequences, producing local embeddings $\mathbf{E}^{\mathrm{loc}}_k\!\in\!\mathbb{R}^{L\times d}$ and a global embedding $\mathbf{E}^{\mathrm{glob}}\!\in\!\mathbb{R}^{L\times d}$. These embeddings provide part-specific and object-level conditioning for the generative process. To integrate these image features into 3D reconstruction, the DiT maintains its own learnable latent tokens that represent the geometry to be generated. Let an object contain $K$ parts, each represented by $N$ latent tokens. We stack them as 
$Z = [Z_1;\dots;Z_K] \in \mathbb{R}^{NK \times C}$, where $Z_i \in \mathbb{R}^{N \times C}$.

During denoising, the model fuses the image embeddings with the latent geometry tokens through alternating \emph{local} and \emph{global} cross-attention blocks. Local blocks query the corresponding part embedding $\mathbf{E}^{\mathrm{loc}}_k$, ensuring each part attends only to its own visual reference, while global blocks query $\mathbf{E}^{\mathrm{glob}}$ to inject whole-object context shared across parts. Within these blocks, latent-to-latent attention is computed using the local map $\mathbf{A}^{\mathrm{local}}_{i}=\mathrm{softmax}\!\big(Z_i Z_i^{\top}/\sqrt{C}\big)\in\mathbb{R}^{N\times N}$ and the global map $\mathbf{A}^{\mathrm{global}}=\mathrm{softmax}\!\big(Z Z^{\top}/\sqrt{C}\big)\in\mathbb{R}^{NK\times NK}$, where $\mathbf{A}$ denotes the attention weights that propagate information within each part and across the full object. To incorporate structural guidance from the VLM priors, we introduce a hierarchical attention mechanism that exchanges information between each parent–child pair. A parent index map $\pi:\{1,\dots,K\}\!\to\!\{-1,1,\dots,K\}$ defines the link hierarchy; $\pi(k)= -1$ for root. Let $\mathcal{P}(u)$ and $\mathcal{C}(u)$ denote the token sets of the parent and child of token $u$, respectively. Child tokens attend only to parent tokens via
$$
A^{c\rightarrow p}_{uv}=
\frac{
\exp\!\big(Z_u Z_v^\top/\sqrt{C}\big)\,\mathbf{1}[v\in\mathcal{P}(u)]
}{
\sum_{v'}\exp\!\big(Z_u Z_{v'}^\top/\sqrt{C}\big)\,\mathbf{1}[v'\in\mathcal{P}(u)]
}.
$$
The updated latent tokens are then computed as $Z' = Z + A^{c\rightarrow p}Z$. Using the updated features $Z'$, parent tokens query their children via
$$
A^{p\rightarrow c}_{uv}=
\frac{
\exp\!\big({Z}_u ({Z}_v)^\top/\sqrt{C}\big)\,\mathbf{1}[v\in\mathcal{C}(u)]
}{
\sum_{v'}\exp\!\big({Z}_u ({Z}_{v'})^\top/\sqrt{C}\big)\,\mathbf{1}[v'\in\mathcal{C}(u)]
}.
$$
We then obtain the final representation as $Z'' = Z' + A^{p\rightarrow c}Z$.
This bidirectional scheme enables the transformer to incorporate local detail, global context, and structural (parent–child) guidance.

\vspace{-10px}
\paragraph{Position Embedding.}
To bind latent sequences to semantic parts and keep the image–part correspondence stable under shuffling, we adopt a dual scheme: (i) a learnable \emph{part embedding} that encodes each part’s \emph{relative} index within its object (0…$K\!-\!1$), and (ii) a learnable \emph{absolute position embedding} that encodes the canonical part identity (e.g., \texttt{link\_0}, \texttt{link\_1}, …). At run time, the data loader supplies absolute indices as \texttt{attention\_kwargs["part\_positions"]}; the transformer then \emph{adds} both embeddings to the hidden states before attention. This ensures that \texttt{link\_0} always receives the same absolute code even if parts are shuffled for augmentation, preserving output order and enabling part-specific features.

\vspace{-10px}
\paragraph{Training Objective and Loss Design.}
We train SPARK using Rectified Flow matching~\cite{liu2022rectified}. For an articulated object with \(K\) parts, the VAE encoder maps each ground-truth part mesh to a latent \(z_{k,0}\in\mathbb{R}^{D}\), while independent base latents \(z_{k,1}\sim\mathcal{N}(0,I)\) are sampled. A shared timestep \(t\in(0,1)\) is drawn for the whole object, and the rectified interpolation is defined as \(x_k(t)=(1-t)z_{k,0}+t z_{k,1}\), forming the stacked latent \(X_t=(1-t)Z_0+tZ_1\). The DiT is conditioned on the global image embedding \(c^{\mathrm{global}}\), part image embeddings \(\{c^{\mathrm{part}}_k\}\), and absolute indices \(\{p_k\}\), ensuring stable input–output alignment. The target velocity field is time-invariant, \(U^\star=Z_0-Z_1\), and the network predicts \(V_\theta(X_t,C,t)\) across all parts. With per-part weights \(\alpha_k\), timestep density \(\rho(t)\), and reweighting \(w(t)\), the Rectified Flow loss becomes
\(\mathcal{L}_{\mathrm{RF}}=\mathbb{E}[\,w(t)\sum_{k=1}^K \alpha_k\|v_\theta(x_k(t),C,t)-u_k^\star\|_2^2\,]
=\mathbb{E}[\,w(t)\|V_\theta(X_t,C,t)-U^\star\|_F^2\,]\).
Finally, absolute \texttt{part\_positions} \([0,\dots,K{-}1]\) preserve semantic ordering under augmentation, and the VAE decoder maps the optimized latents to per-part meshes that are assembled into the articulated object.
\vspace{-10px}
\paragraph{Texture Generation.} 
After obtaining the articulated mesh, we apply textures to each kinematic part using Meshy~\cite{meshy-ai}, guided by the corresponding per-part reference images. Since the generated part-level meshes may vary in scale or position, we further employ the Iterative Closest Point (ICP) algorithm \cite{rusinkiewicz2001efficient} to align the textured meshes consistently with the input image.

\vspace{-3px}
\subsection{Joint Optimization}
\vspace{-3px}
Generating URDF parameters from images or 3D shapes can be achieved with end-to-end models, but such data-driven approaches often struggle to accurately estimate joint parameters due to the lack of kinematic guidance. To address this limitation, we apply a feature-injection strategy in the VLM prediction module to refine discrete joint parameters, such as the joint axis $\mathbf{u}_{j}^{\text{axis}}$ and joint type $\mathbf{u}_{j}^{\text{type}}$. More specifically, we use the coarse URDF parameters together with the input image as conditions for the VLM to re-predict the $\mathbf{u}_{j}^{\text{axis}}$ and $\mathbf{u}_{j}^{\text{type}}$.

For continuous parameter optimization, we incorporate differentiable forward kinematics and differentiable rendering to recover accurate continuous joint parameters, including the joint origin $\mathbf{u}_{j}^{\text{origin}}$ and the joint angle $\Delta\theta \in \mathbb{R}$, guided by the generated open-state reference image $I_{\text{open}}$. Let $\boldsymbol{\xi} = (\Delta\mathbf{t},\, \Delta\theta)$ denote the learnable joint parameters, where $\Delta\mathbf{t}\!\in\!\mathbb{R}^3$ represents the joint origin in the parent frame and $\Delta\theta\!\in\!\mathbb{R}$ is the rotation angle about the predefined axis. These parameters define a rigid-body rotation in $\mathrm{SO}(3)$ that determines the local motion of the child link. Starting from the closed-state articulated object $M^0$, the differentiable forward kinematics function $G(\cdot)$ is applied to obtain the transformed object 
$M^{\mathrm{t}} = G(M^0,\, \Delta\mathbf{t},\, \Delta\theta)$. 
Given a fixed camera $\mathcal{C}$, we employ a differentiable renderer $\mathcal{R}$ to generate a soft silhouette  $I_{\text{sil}} = \mathcal{R}(\mathcal{C},\, M^{\mathrm{t}})$ 
from the transformed object $M^{\mathrm{t}}$. To supervise the optimization, our objective is to: 
\begin{equation}
\min_{\boldsymbol{\xi}}
\ \ \mathcal{L}_{\mathrm{total}}
= \mathcal{L}_{\mathrm{pixel}}\!\left(I_{\text{sil}},\, I_{\text{open}}\right)
+ \mathcal{L}_{\mathrm{reg}}(\boldsymbol{\xi}).
\label{eq:joint-opt-total}
\end{equation}
Here, $\mathcal{L}_{\mathrm{pixel}}$ denotes a pixel-level silhouette loss that measures the discrepancy between the rendered silhouette $I_{\text{sil}}$ and the open-state reference image $I_{\text{open}}$, promoting accurate region and boundary alignment. Specifically, $\mathcal{L}_{\mathrm{pixel}}$ combines a region loss, $\mathcal{L}_{\mathrm{region}} = 1 - \frac{2\langle I_{\text{sil}}, I_{\text{open}} \rangle}{\|I_{\text{sil}}\| + \|I_{\text{open}}\|}$, which emphasizes region overlap and contour consistency, and a gradient-based edge loss, $\mathcal{L}_{\mathrm{edge}} = \|\,|\nabla I_{\text{sil}}| - |\nabla I_{\text{open}}|\,\|$, which preserves edge sharpness and boundary similarity. Moreover, we explore the regularization term $\mathcal{L}_{\mathrm{reg}}$ that constrains the joint translation $\Delta\mathbf{t}$ and rotation $\Delta\theta$ to remain close to the initial position, preventing instability and unrealistic motion:
\begin{equation}
\mathcal{L}_{\mathrm{reg}} = \lambda_t\,\|\Delta\mathbf{t}\|_2^2 + \lambda_\theta\,\|\Delta\theta\|_2^2.
\end{equation}Here, $\lambda_t$ and $\lambda_\theta$ are weighting coefficients that balance the regularization strength between translational and rotational offsets, ensuring numerical stability during optimization.

\vspace{-3px}
\subsection{Data Curation and Augmentation}
\vspace{-3px}
\begin{figure*}[t]
    \centering
    \includegraphics[width=0.9\textwidth]{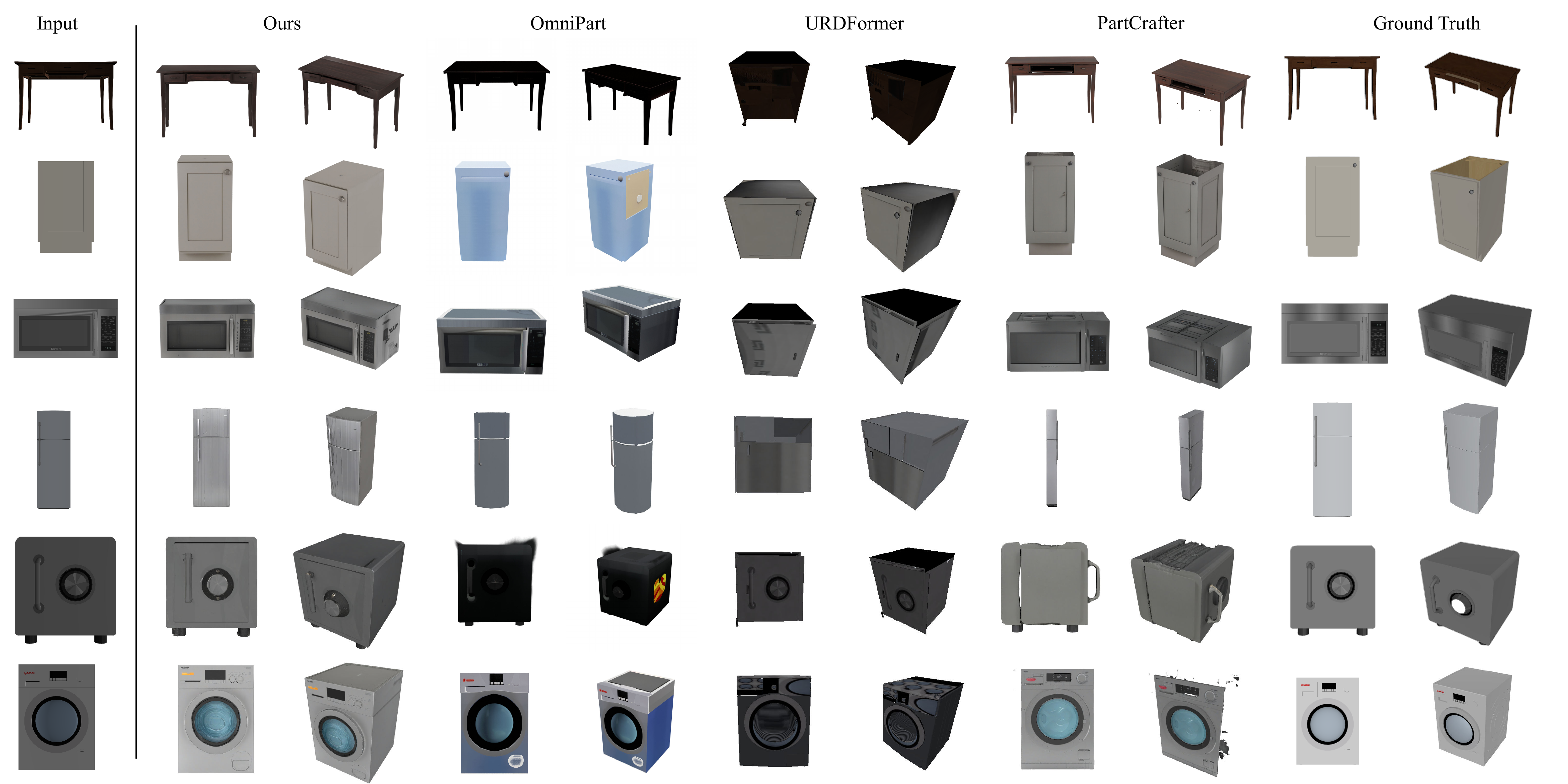}
    \vspace{-5pt}
    \caption{
        \textbf{Qualitative Comparison on Shape Reconstruction.} We compare our results with OmniPart~\cite{yang2025omnipart}, PartCrafter~\cite{lin2025partcrafter}, and URDFormer~\cite{chen2024urdformer}. Our method fulfills accurate, high-fidelity articulated object shape reconstruction. 
    }
    \label{fig:qua_mesh}
    \vspace{-5pt}
\end{figure*}

\begin{figure*}[t]
    \centering
    \includegraphics[width=0.9\textwidth]{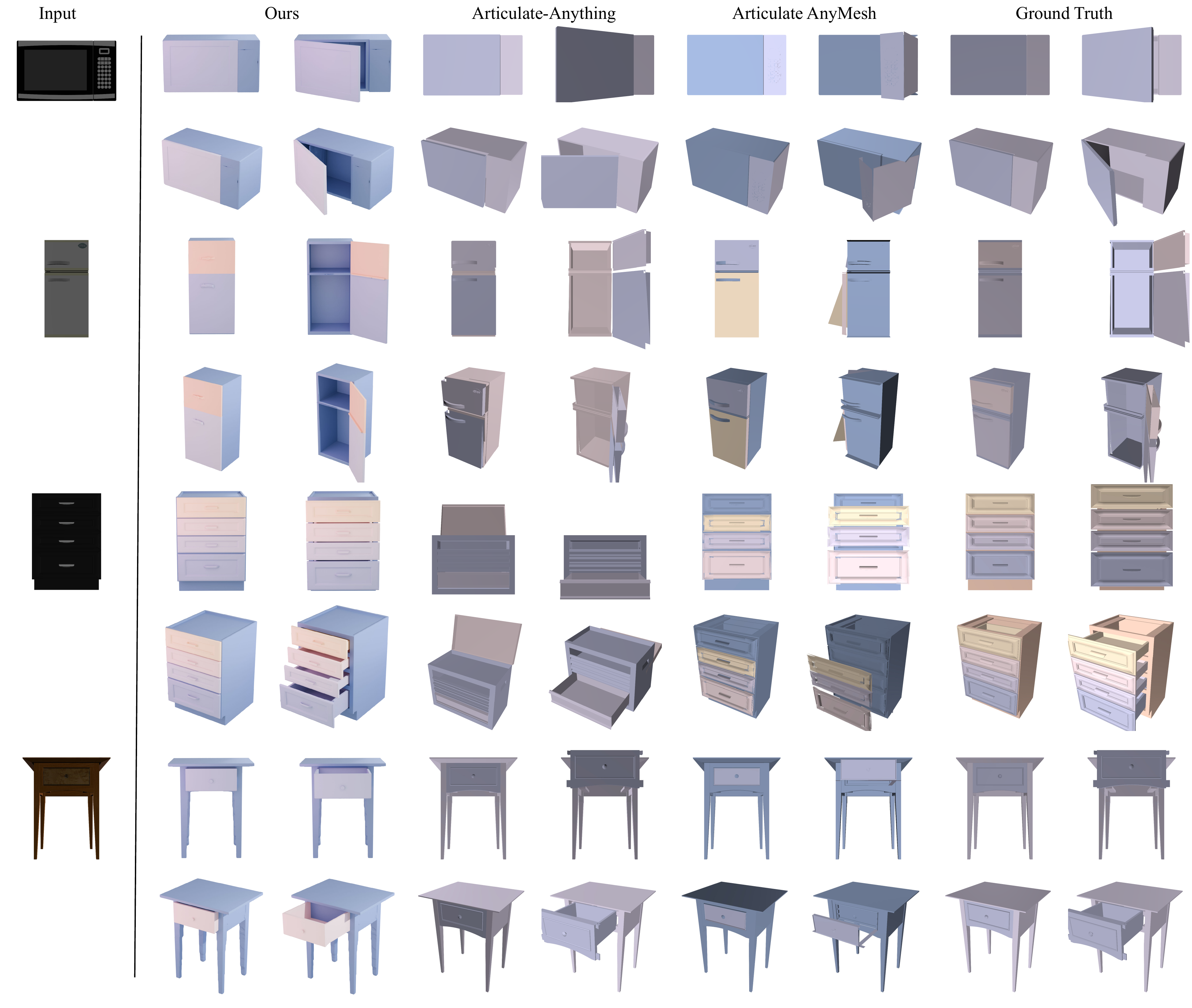}
    \vspace{-5pt}
    \caption{
        \textbf{Qualitative Comparison on URDF Estimation.} We compare our results with Articulate-Anything~\cite{le2024articulate}, Articulate-AnyMesh~\cite{qiu2025articulate}. 
        The closed-state results are reconstructed or retrieved meshes, while the open-state configurations are obtained through kinematic transformations using the estimated URDF parameters. Our method achieves more accurate and physically consistent URDF estimation, leading to realistic articulation behavior.
    }
    \label{fig:qua_urdf}
    \vspace{-5pt}
\end{figure*}

SPARK is trained on the open-source PartNet-Mobility dataset~\cite{Xiang_2020_SAPIEN}, which contains $2,347$ articulated objects across $46$ categories in a consistent format. This dataset provides paired meshes and rendered images, but only in a single canonical articulation state. In contrast, general 3D generation datasets such as Objaverse~\cite{objaverse} include objects captured in diverse articulation states—for example, drawers that are partially opened or laptops at various folding angles.  In addition, we observe that some assets are over-segmented, where a single kinematic link is divided into multiple mesh pieces. To improve data quality and articulation diversity, we perform targeted data curation and augmentation. Specifically, we merge over-segmented meshes according to the URDF link associations provided in the annotations to obtain a single consolidated mesh per link. Moreover, to increase diversity, we augment articulation states by sampling joint configurations across feasible motion ranges, generating multiple canonical poses per object and capturing realistic articulation variations (e.g., partially opened drawers or unfolded laptops).

%% file: sec/4_experiments.tex
\section{Experiments}
\label{sec:exp}
\vspace{-3px}

\vspace{-3px}
\paragraph{Implementation Details.}
We implement the differentiable forward kinematics module in PyTorch~\cite{paszke2019pytorch} and use PyTorch3D~\cite{ravi2020accelerating} for differentiable rendering. For VLM-guided structural reasoning, we employ GPT-4o to extract part labels and joint metadata from the input image, and use Gemini 2.5 Flash Image (Nano Banana) to synthesize both the part images and the open-state image. Our training data are built upon the PartNet-Mobility dataset~\cite{Xiang_2020_SAPIEN}, augmented with three articulation states: open, closed, and half-open. To ensure high-quality samples, we repair non-watertight meshes, which would otherwise introduce surface artifacts, by voxelizing each mesh and re-extracting its surface at a resolution of 200. We train our models on four NVIDIA H100 GPUs with a batch size of 48 and a learning rate of $1\times 10^{-4}$ for 1,000 epochs, requiring around 60 hours.

\vspace{-12px}
\paragraph{Baseline.}
We divide our evaluation into articulated object shape reconstruction and URDF estimation. For shape reconstruction, we compare against state-of-the-art part-aware generation methods, including PartCrafter~\cite{lin2025partcrafter} and OmniPart~\cite{yang2025omnipart}, which generate 3D objects with functionally components, as well as URDFormer~\cite{chen2024urdformer}, which assembles articulated objects using structural guidance. For URDF estimation, we benchmark against Articulate Anything~\cite{le2024articulate} and Articulate Anymesh~\cite{qiu2025articulate}, aiming to recover URDF from a single image and a 3D mesh, respectively. 

\begin{table}[t]
\centering
\caption{\textbf{Quantitative Shape Reconstruction Comparison.} 
We report F-score to measure reconstruction accuracy, Chamfer Distance (CD) for geometric fidelity.}
\setlength\tabcolsep{5pt} 
\label{tab:quan_mesh}
\small{
\begin{tabular}{p{1.05in}ccc} 
\hline
\textbf{Methods} & CD$\downarrow$ & F-Score@0.1$\uparrow$ & F-Score@0.5$\uparrow$ \\
\hline
PartCrafter~\cite{lin2025partcrafter} & \cellcolor{Palette4}{0.4342} & \cellcolor{Palette4}{0.3600} & \cellcolor{Palette4}{0.8840} \\
OmniPart~\cite{yang2025omnipart}      & 0.4971 & 0.1928 & 0.8469 \\
URDFormer~\cite{chen2024urdformer}    & 1.0556 & 0.0438 & 0.1762 \\
\hline
Ours & \cellcolor{Palette5}{0.3959} & \cellcolor{Palette5}{0.4214} & \cellcolor{Palette5}{0.8934} \\
\hline
\end{tabular}
\vspace{-5px}
}
\end{table}

\subsection{Quantitative Evaluation}
\label{sec:quan}

\begin{table}[t]
\centering
\caption{\textbf{Quantitative URDF Parameter Estimation Comparison.} 
We evaluate articulated object URDF parameter estimation using AxisErr, PivotErr, and TypeErr, which measure joint axis deviation, joint pivot offset, and joint type misclassification.}

\setlength\tabcolsep{5pt}
\label{tab:quan_urdf}
\small{
\begin{tabular}{p{1.25in}ccc}
\hline
\textbf{Methods} & AxisErr$\downarrow$ & PivotErr$\downarrow$ & TypeErr$\downarrow$ \\
\hline
Articulate-Anything \cite{le2024articulate} 
& \cellcolor{Palette4}{0.5491} 
& \cellcolor{Palette4}{0.3529} 
& \cellcolor{Palette4}{0.2500} \\
Articulate Anymesh \cite{qiu2025articulate}       
& 1.1834 & 0.9162 & 0.7000 \\
\hline
Ours                                        
& \cellcolor{Palette5}{0.1577} 
& \cellcolor{Palette5}{0.1653} 
& \cellcolor{Palette5}{0.0500} \\
\hline
\end{tabular}
\vspace{-5px}
}
\end{table}

To evaluate reconstruction quality, we use a test set of 50 images covering 25 articulated object categories from GAPartNet~\cite{geng2023gapartnet}. For each method, we reconstruct meshes from the input images and compare them with the ground-truth geometry. Following prior work~\cite{lin2025partcrafter, yang2025omnipart}, we adopt two standard metrics: Chamfer Distance for geometric fidelity and F-Score for point-level alignment. We report F-Score@0.1 (strict matching) and F-Score@0.5 (coarse matching). For URDF parameter estimation, we compare against Articulate-Anything~\cite{le2024articulate} and Articulate AnyMesh~\cite{qiu2025articulate}, which predict joint axes, pivots, and type from images and meshes, respectively. Following prior work in articulated object understanding~\cite{chen2025freeart3d}, we quantify accuracy using AxisErr (angular deviation), PivotErr (positional error), and TypeErr (type classification).

\vspace{-10px}
\paragraph{Articulated Object Shape Reconstruction.}   As shown in Tab.~\ref{tab:quan_mesh}, our method outperforms all baselines across both metrics. Although PartCrafter and OmniPart achieve comparable performance at the loose threshold (F-Score@0.5), our approach significantly surpasses them at the strict threshold (F-Score@0.1), demonstrating superior fine-grained geometric recovery. URDFormer underperforms in both F-Score and Chamfer Distance because its bounding-box–driven template retrieval strategy can misidentify part types, leading to suboptimal geometric alignment and reduced reconstruction accuracy. In contrast, our method leverages part-level image and structure guidance from VLM priors, enabling accurate geometry generation with consistent topology and improved visual quality.

\vspace{-10px}
\paragraph{URDF Parameter Estimation Comparison.}
As shown in Tab.~\ref{tab:quan_urdf}, our method achieves substantial improvements in URDF parameter estimation across all metrics. The performance gap is particularly pronounced for continuous parameters, where our optimization component enables much more accurate recovery of joint axes and joint origins compared with Articulate-Anything and Articulate AnyMesh. Articulate-Anything relies on vision-language program synthesis without any geometry-based refinement, causing its predicted joint axes and pivots to deviate significantly from the true kinematic configuration. Articulate AnyMesh, in turn, depends heavily on noisy connecting-area geometry and heuristic hinge-point selection, which makes its joint axis and origin estimates sensitive to segmentation errors and mesh artifacts, as indicated in Fig. \ref{fig:qua_urdf}. 

\vspace{-5px}
\subsection{Qualitative Evaluation}

Fig.~\ref{fig:qua_mesh} presents qualitative comparisons with baseline methods on the same test set described in Section~\ref{sec:quan}. PartCrafter often produces floating or disconnected components and may miss key geometric elements, such as table drawers or cabinet top surfaces. URDFormer, which relies on template retrieval, can generate mismatched part types, for instance, producing small cabinet-like ledges in reconstructed tables. OmniPart first segments the image and then reconstructs each part from its bounding box, making it highly sensitive to segmentation noise and box misalignment; as a result, it often yields distorted shapes in occluded or unseen regions, such as cabinet interiors or doors.

\begin{table}[t]
\centering
\caption{\textbf{Mesh Reconstruction Ablation.} Ablation on mesh reconstruction quality.}
\setlength\tabcolsep{4pt} 
\label{tab:ab_mesh}
\small{
\begin{tabular}{p{0.95in}ccc} 
\hline
\textbf{Variants}  & CD$\downarrow$ & F-Score@0.1$\uparrow$ & F-Score@0.5$\uparrow$ \\
\hline
w/o Part Guidance 
& 0.4284 & 0.3755 & 0.8725 \\
w/o Data Aug.     
& 0.4200     & 0.3675     & 0.8883     \\
\hline
Ours              & \cellcolor{Palette5}{0.3959} 
                 & \cellcolor{Palette5}{0.4214} 
                 & \cellcolor{Palette5}{0.8934} \\
\hline
\end{tabular}
\vspace{-6px}
}
\end{table}

\begin{table}[t]
\centering
\caption{\textbf{URDF Parameter Estimation Ablation.} We evaluate our joint optimization using the same metrics to assess URDF parameter estimation accuracy.}

\setlength\tabcolsep{5pt}
\label{tab:ab_urdf}
\small{
\begin{tabular}{p{1.25in}ccc}
\hline
\textbf{Variants} & AxisErr$\downarrow$ & PivotErr$\downarrow$ & TypeErr$\downarrow$\\
\hline
w/o Joint Optimization      
& 0.3148 & 0.2388 & 0.2000 \\
\hline
Ours  
& \cellcolor{Palette5}{0.1577} & \cellcolor{Palette5}{0.1653}  & \cellcolor{Palette5}{0.0500} \\
\hline
\end{tabular}
\vspace{-5px}
}
\end{table}

\begin{figure*}[t]
    \centering
        \includegraphics[width=\textwidth]{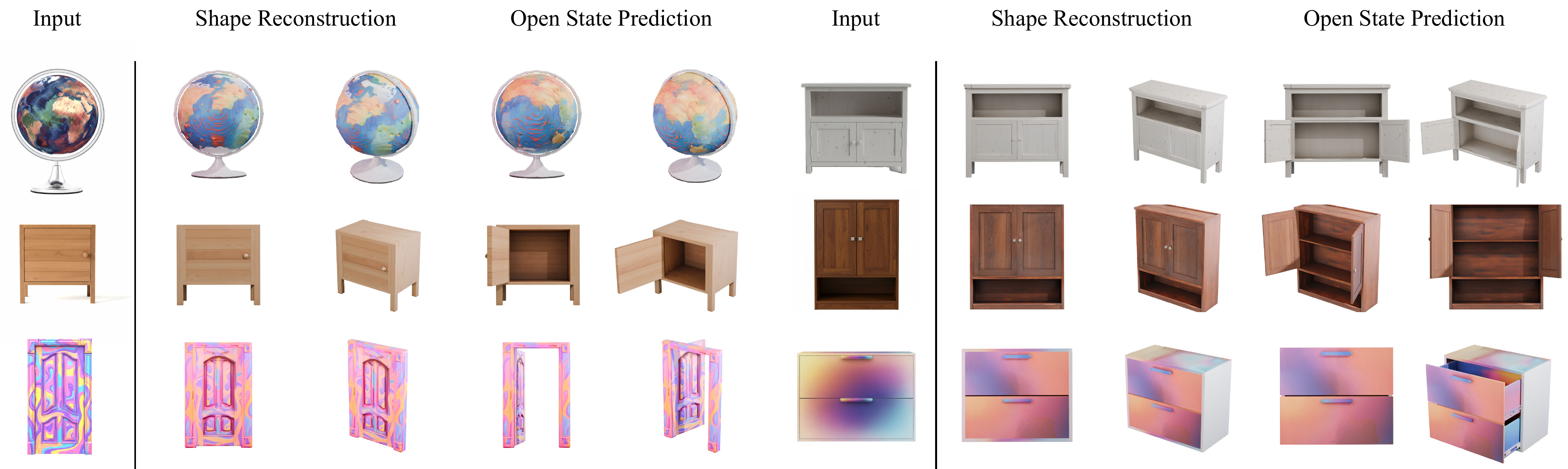}
        \vspace{-0.3in}
    \caption{\small{\textbf{In-the-wild image results.}
Additional examples of shape reconstruction and open-state prediction on in-the-wild images.}}\label{fig:sub_inthewild}
\label{fig:inthewild}
\end{figure*}

For URDF parameter estimation, retrieval-based methods such as Articulate-Anything and Articulate AnyMesh start by retrieving a closed-state object with a similar overall shape. While this can produce roughly plausible geometry, it frequently retrieves incorrect part types, such as mismatched drawers in Fig.~\ref{fig:qua_urdf}. These methods also apply predicted URDF parameters directly to segmented parts, making them vulnerable to segmentation errors that lead to incorrect kinematic motions. For example, in the fridge case, Articulate-Anything misidentifies the side surface as the door, causing inconsistent opening motions, while Articulate AnyMesh may confuse the outer drawer surface with the internal drawer. In contrast, our optimization component yields substantially more accurate URDF parameters and coherent, physically plausible articulated motions. Additional in-the-wild results are provided in Fig.\ref{fig:inthewild}.

\vspace{-3px}
\subsection{Ablation Study}
\begin{figure}{}
    \vspace{-10pt}
    \centering
    \includegraphics[width=0.9\linewidth]{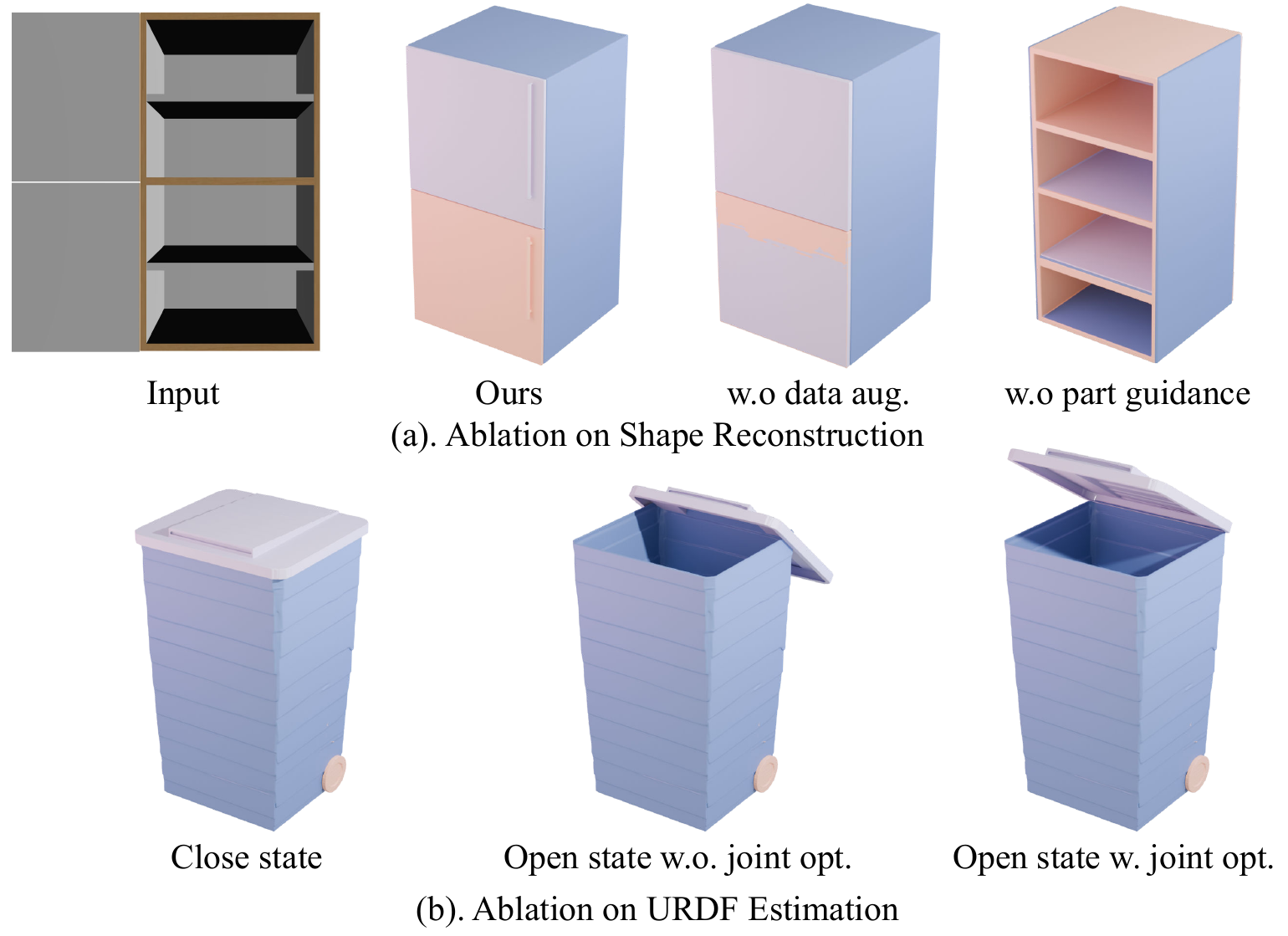}
    \caption{
        \textbf{Ablation.} We conduct an ablation study on data augmentation, part guidance, and joint optimization.
    }
    \vspace{-12pt}
    \label{fig:ab}
\end{figure}

\vspace{-3px}
We conduct an ablation study on the key components of our framework: part guidance, data augmentation, and joint optimization, to evaluate their contributions to mesh reconstruction and URDF parameter estimation. VLM-driven part guidance provides strong visual and structural priors. As shown in Fig.~\ref{fig:ab}, removing part guidance leads to missing doors in the cabinet. Without data augmentation, the model overfits to limited states and may confuse visually similar parts (e.g., misidentifying two cabinet doors). Incorporating both VLM-derived guidance and data augmentation yields cleaner part boundaries, more consistent decomposition, and higher-quality reconstructions. As reported in Table~\ref{tab:ab_mesh}, these components together achieve the best geometric accuracy. Joint optimization further refines URDF parameters, improving joint-type prediction, joint-axis orientation, and joint pivot estimation (Table~\ref{tab:ab_urdf}). Removing optimization leads to misaligned or drifting moving parts, such as shifted pivot points (Fig. \ref{fig:ab}). 

\setlength{\columnsep}{6pt}

\vspace{-5px}
\subsection{Applications}

\begin{figure}{}
    \centering
    \includegraphics[width=\linewidth]{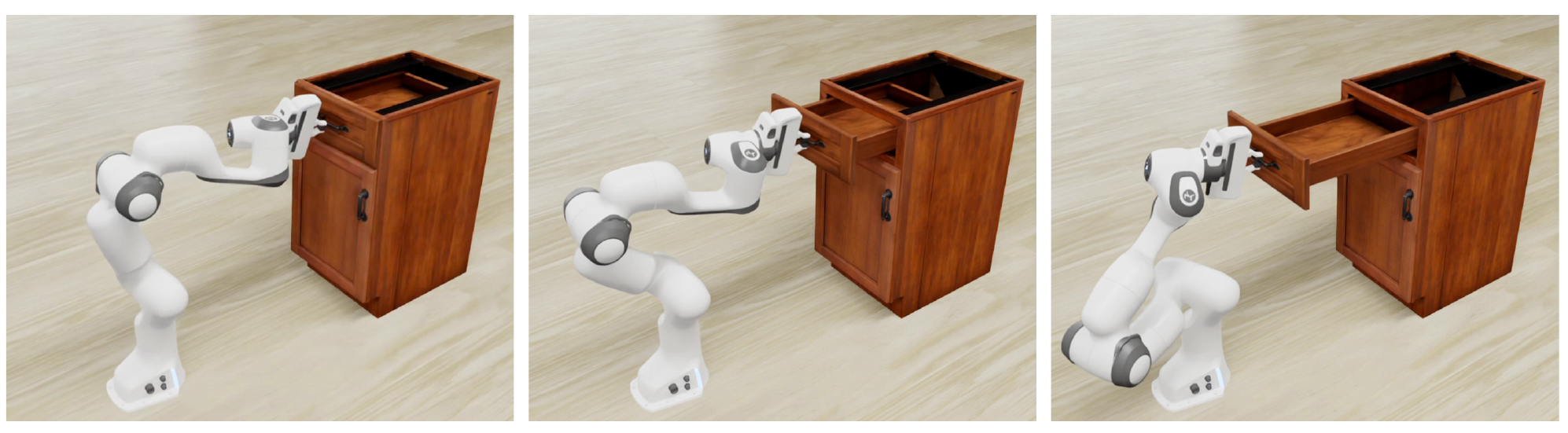}
    \caption{
        \textbf{Robot Learning.} We use the synthesized drawer to train a robot in Isaac Sim \cite{makoviychuk2021isaac} to open a drawer.
    }
    \vspace{-12pt}
    \label{fig:robot}
\end{figure}

Our method generates high-quality articulated objects with accurate geometry and URDF parameters, enabling robot learning of tasks such as door opening and drawer pulling. We demonstrate this through a robot drawer-opening task on the generated drawer (Fig.~\ref{fig:robot}).

\setlength{\columnsep}{6pt}

%% file: sec/5_conclusion.tex
\section{Conclusion}
\label{sec:conclusion}
\vspace{-5px}
We presented \textbf{SPARK}, a novel framework for reconstructing simulation-ready articulated objects from a single image, guided by VLM-derived part images and structural cues. By combining vision–language priors, a diffusion transformer with hierarchical attention, and a differentiable joint optimization module, SPARK recovers high-quality part-level geometry and accurate articulation parameters. In future work, we plan to extend SPARK to more complex kinematic structures beyond simple revolute and prismatic joints, including multi-DOF joints, compound mechanisms, and closed-chain linkages commonly found in real-world appliances and tools.

%% file: sec/arxiv_sup.tex
\section{Network Architecture Details}
\label{sec:network-appendix}

In this section, we provide additional details of the network backbone used by SPARK. Our model follows a part-aware Diffusion Transformer (DiT) design with dual-image conditioning and hierarchical attention over kinematic links, built on top of a Variational Autoencoder (VAE).

\subsection{Latent Representation and Image Encoders}

Given an articulated object with $K$ semantic parts, we represent each part $k$ by a point-based surface encoding and a reference image:
\begin{itemize}
    \item \textbf{Part surfaces}. Each ground-truth part mesh is sampled into $P$ oriented surface points, yielding $\mathbf{S}_k \in \mathbb{R}^{P \times 6}$ (3D position + normal). A shared VAE encoder maps $\mathbf{S}_k$ to a latent sequence
    \[
        \mathbf{z}_{k,0} \in \mathbb{R}^{T \times D_{\text{lat}}},
    \]
    where $T$ is the number of latent tokens per part and $D_{\text{lat}}$ is the latent dimension.
    \item \textbf{Part and global images}. For conditioning, we use a per-part reference image $r_k$ and a single global input image $I_0$. The global image is replicated $K$ times so each part sees the same holistic context.
\end{itemize}

Both $r_k$ and $I_0$ are encoded by a shared DINOv2 image encoder into token sequences
\[
    E^{\text{loc}}_k \in \mathbb{R}^{L \times d_{\text{img}}}, 
    \qquad
    E^{\text{glob}} \in \mathbb{R}^{L \times d_{\text{img}}},
\]
where $L$ is the number of visual tokens and $d_{\text{img}}$ is the image feature dimension. The global embedding $E^{\text{glob}}$ is broadcast across parts, while $E^{\text{loc}}_k$ carries part-specific appearance cues.

We stack the $K$ per-part latent sequences along the batch axis to obtain
\[
    \mathbf{Z}_0 = 
    \begin{bmatrix}
        \mathbf{z}_{1,0} \\
        \vdots \\
        \mathbf{z}_{K,0}
    \end{bmatrix}
    \in \mathbb{R}^{(K T) \times D_{\text{lat}}},
\]
which is the clean latent representation used in the diffusion process.

\subsection{Diffusion Transformer Backbone}

The generative backbone is a DiT with $L_{\text{DiT}}$ transformer blocks and hidden dimension $D$ (we use $D \gg D_{\text{lat}}$). Before entering the transformer, we project the VAE latents and inject a timestep token:
\begin{align}
    \mathbf{H} &= \mathrm{Proj}(\mathbf{Z}_t) \in \mathbb{R}^{(K T) \times D}, \\
    \mathbf{h}_{\text{time}} &= \phi_{\text{time}}(t) \in \mathbb{R}^{1 \times D}, \\
    \tilde{\mathbf{H}} &= \big[\, \mathbf{h}_{\text{time}}; \mathbf{H} \,\big] \in \mathbb{R}^{(K T + 1) \times D},
\end{align}
where $\mathbf{Z}_t$ is the noisy latent at time $t$ (defined in Sec.~\ref{sec:network-appendix:rf}), $\phi_{\text{time}}$ is a sinusoidal+MLP timestep embedding, and $[\,\cdot;\cdot\,]$ denotes concatenation along the token dimension.

\paragraph{Part and position embeddings.}
To make the model part-aware and robust to part shuffling, we add two learnable embeddings per part:
\begin{itemize}
    \item A \emph{relative part embedding} $e^{\text{rel}}(k)$ that encodes the index of part $k$ within its object, $k \in \{0,\dots,K{-}1\}$.
    \item An \emph{absolute position embedding} $e^{\text{abs}}(p_k)$ that encodes the canonical identity of the corresponding link (e.g., \texttt{link\_0}, \texttt{link\_1}, \dots).
\end{itemize}
During data loading, we maintain an array of absolute indices $p_k$ that is preserved under shuffling, so each semantic link always receives a consistent absolute embedding even when the parts are randomly permuted for augmentation. For each token that belongs to part $k$, we add both embeddings to the hidden state:
\begin{equation}
    \tilde{\mathbf{H}} \leftarrow \tilde{\mathbf{H}} 
    + \mathrm{Broadcast}\big(e^{\text{rel}}(k)\big)
    + \mathrm{Broadcast}\big(e^{\text{abs}}(p_k)\big),
\end{equation}
where $\mathrm{Broadcast}(\cdot)$ expands a per-part vector to all tokens of that part.

\subsection{Multi-Level Attention and Image Conditioning}

We interleave three types of attention to combine part-level geometry, local appearance, and global context:

\paragraph{Self-attention.}
Each transformer block first applies self-attention with rotary positional encoding over the latent tokens:
\begin{equation}
    \mathbf{A}^{\text{self}} = \mathrm{softmax}\big(\mathbf{Q}\mathbf{K}^\top / \sqrt{D}\big),
    \qquad
    \mathbf{H}' = \mathbf{H} + \mathbf{A}^{\text{self}}\mathbf{V},
\end{equation}
where $\mathbf{Q},\mathbf{K},\mathbf{V}$ are the usual query, key, and value projections of $\tilde{\mathbf{H}}$.

\paragraph{Global cross-attention.}
In a subset of layers indexed by $\mathcal{L}_{\text{glob}}$, we use cross-attention on the global image embedding $E^{\text{glob}}$ to inject holistic object context shared across all parts:
\begin{equation}
    \mathbf{A}^{\text{glob}} = \mathrm{softmax}\big(\mathbf{Q}\mathbf{K}^{\text{glob}\,\top} / \sqrt{D}\big),
    \qquad
    \mathbf{H}'' = \mathbf{H}' + \mathbf{A}^{\text{glob}}\mathbf{V}^{\text{glob}},
\end{equation}
where $\mathbf{K}^{\text{glob}},\mathbf{V}^{\text{glob}}$ are projections of $E^{\text{glob}}$.

\paragraph{Local cross-attention.}
In the remaining layers, indexed by $\mathcal{L}_{\text{loc}}$, we instead use cross-attention to the part-specific embeddings $E^{\text{loc}}_k$. Tokens belonging to part $k$ only attend to the visual features of that part:
\begin{equation}
    \mathbf{A}^{\text{loc}}_k 
    = \mathrm{softmax}\big(\mathbf{Q}_k\mathbf{K}^{\text{loc}\,\top}_k / \sqrt{D}\big),
    \qquad
    \mathbf{H}''_k = \mathbf{H}'_k + \mathbf{A}^{\text{loc}}_k\mathbf{V}^{\text{loc}}_k.
\end{equation}
Alternating global and local layers encourages the model to respect both global shape consistency and part-level details driven by the per-part images.

\subsection{Hierarchy Attention Implementation}
\label{sec:network-appendix:hier-attn}

To explicitly encode the predicted kinematic tree, we adopt a hierarchy attention module operating on parent--child link pairs. Let $\pi: \{1,\dots,K\} \to \{-1,1,\dots,K\}$ denote the parent index map over parts, where $\pi(k)=-1$ for the root. We share this map across all tokens of a part.

We first define the sets of parent and child tokens for a given token $u$:
\[
    P(u) = \{ v \mid \text{token $v$ belongs to the parent part of $u$} \}, 
\]
\[
    C(u) = \{ v \mid \text{token $v$ belongs to a child part of $u$} \}.
\]

\paragraph{Child-to-parent attention.}
Given latent tokens $\mathbf{Z}$, we compute a child-to-parent attention matrix
\begin{equation}
    A^{c \rightarrow p}_{uv}
    =
    \frac{
        \exp\big( \mathbf{z}_u \mathbf{z}_v^\top / \sqrt{C} \big)\,\mathbf{1}[v \in P(u)]
    }{
        \sum\limits_{v'} \exp\big( \mathbf{z}_u \mathbf{z}_{v'}^\top / \sqrt{C} \big)\,\mathbf{1}[v' \in P(u)]
    },
\end{equation}
and update the latent tokens via
\begin{equation}
    \mathbf{Z}' = \mathbf{Z} + A^{c \rightarrow p}\mathbf{Z}.
\end{equation}
This step lets each child part aggregate structural context from its parent.

\paragraph{Parent-to-child attention.}
We then allow parents to read back from their children using
\begin{equation}
    A^{p \rightarrow c}_{uv}
    =
    \frac{
        \exp\big( \mathbf{z}_u \mathbf{z}_v^\top / \sqrt{C} \big)\,\mathbf{1}[v \in C(u)]
    }{
        \sum\limits_{v'} \exp\big( \mathbf{z}_u \mathbf{z}_{v'}^\top / \sqrt{C} \big)\,\mathbf{1}[v' \in C(u)]
   },
\end{equation}
and obtain the final hierarchy-aware representation
\begin{equation}
    \mathbf{Z}'' = \mathbf{Z}' + A^{p \rightarrow c}\mathbf{Z}'.
\end{equation}

In practice, we implement this module as a separate transformer block that takes the current hidden states and the batched parent indices as input. To avoid cross-sample contamination, we offset parent indices per object and explicitly validate that no part attends to parents outside its own object. For efficiency, we compute aggregated parent/child features using scatter-add operations rather than explicit Python loops.

\subsection{Rectified Flow Training Objective}
\label{sec:network-appendix:rf}

We train the DiT backbone using Rectified Flow matching. For each object, we draw an independent base latent per part,
\[
    \mathbf{z}_{k,1} \sim \mathcal{N}(0, \mathbf{I}),
\]
and define the clean and base stacks
\[
    \mathbf{Z}_0 =
    \begin{bmatrix}
        \mathbf{z}_{1,0} \\
        \vdots \\
        \mathbf{z}_{K,0}
    \end{bmatrix},
    \qquad
    \mathbf{Z}_1 =
    \begin{bmatrix}
        \mathbf{z}_{1,1} \\
        \vdots \\
        \mathbf{z}_{K,1}
    \end{bmatrix}.
\]
A shared timestep $t \in (0,1)$ is sampled per object from a non-uniform logit-normal density $\rho(t)$ that emphasizes informative ranges. The interpolated latent is
\begin{equation}
    \mathbf{X}_t = (1 - t)\mathbf{Z}_0 + t \mathbf{Z}_1,
\end{equation}
and the target velocity field is time-invariant,
\begin{equation}
    \mathbf{U}^\star = \mathbf{Z}_0 - \mathbf{Z}_1.
\end{equation}

Let $C$ collect all conditioning signals: the global image embedding $c_{\text{global}}$, the per-part image embeddings $\{c^{\text{part}}_k\}$, and the absolute indices $\{p_k\}$. The DiT predicts a velocity field $V_\theta(\mathbf{X}_t, C, t)$ over all tokens. With per-part weights $\alpha_k$, timestep density $\rho(t)$, and a scalar reweighting function $w(t)$, we optimize
\begin{equation}
\begin{aligned}
\mathcal{L}_{\text{RF}}
&=
\mathbb{E}_{t,\,\mathbf{Z}_0,\,\mathbf{Z}_1,\,C}
\Big[
    w(t)
    \sum_{k=1}^K
    \alpha_k
    \big\|
        v_\theta(\mathbf{x}_k(t), C, t)
        -
        \mathbf{u}_k^\star
    \big\|_2^2
\Big] \\
&=
\mathbb{E}
\Big[
    w(t)
    \big\|
        V_\theta(\mathbf{X}_t, C, t) - \mathbf{U}^\star
    \big\|_F^2
\Big].
\end{aligned}
\end{equation}

Here, we use a reverse-velocity parameterization consistent with our sampler, and share $t$ across all parts of the same object to keep the noise level aligned within an articulated asset.

\subsection{Classifier-Free Guidance}

We adopt classifier-free guidance at the object level. During training, with probability $p_{\text{cfg}}$ we drop all image conditions for an entire object and replace both $E^{\text{loc}}_k$ and $E^{\text{glob}}$ by learned ``null'' embeddings, yielding an unconditional branch. The DiT is thus trained on a mixture of conditional and unconditional samples.

At inference time, we evaluate the network twice for each diffusion step: once with all conditions dropped ($V_\theta^{\text{uncond}}$) and once with full conditioning ($V_\theta^{\text{cond}}$). The guided prediction is
\begin{equation}
    V_\theta^{\text{guid}} 
    =
    V_\theta^{\text{uncond}}
    +
    s_{\text{cfg}}
    \big(
        V_\theta^{\text{cond}}
        -
        V_\theta^{\text{uncond}}
    \big),
\end{equation}
where $s_{\text{cfg}}$ is the guidance scale. This formulation lets us trade off fidelity to the input image against sample diversity while preserving multi-part consistency.

\section{Kinematic-Part Mesh Merging}
\label{sec:merge-to-glb}

We unify the raw PartNet-Mobility meshes into a single canonical GLB file per object, where each URDF link corresponds to exactly one mesh. In the original dataset, a single kinematic link may reference multiple OBJ files (e.g., different materials or subcomponents). For our purposes, a \emph{kinematic part} is defined at the link level, so each link must appear as one rigid mesh that moves as a unit under the URDF articulation. We therefore merge all OBJs associated with the same link into a single geometry and export them.

\subsection{URDF-Driven Link Grouping}

For each object, we start from the original \texttt{mobility.urdf}. We parse all \texttt{<link>} elements and collect their associated mesh filenames from the \texttt{<visual>} blocks:
\begin{itemize}
    \item We treat every non-base link (i.e., links whose name is not \texttt{base}) as a candidate kinematic part.
    \item For each such link, we traverse all \texttt{<visual>} elements and extract the \texttt{filename} attribute of the nested \texttt{<mesh>} tag.
    \item The original paths typically resemble \texttt{textured\_objs/original-50.obj}; we reduce them to basenames (e.g., \texttt{original-50.obj}) and assume the corresponding geometry resides in the \texttt{textured\_objs/} folder.
\end{itemize}

This yields a mapping
\[
    \mathcal{G}: \text{link name} \;\mapsto\; \{\text{OBJ filenames}\},
\]
which defines how raw meshes should be grouped into kinematic parts. If a link has no valid mesh entries in the URDF, it is skipped.

\subsection{Per-Link Geometry Cleaning and Merging}

We load and merge meshes on a per-link basis using \texttt{trimesh}. For each link $\ell$ with mesh file list $\mathcal{G}(\ell)$:
\begin{enumerate}
    \item We attempt to load each OBJ file from \texttt{textured\_objs/}. Files that cannot be loaded (missing or malformed) are skipped with a warning.
    \item Each successful load is converted to a pure-geometry \texttt{Trimesh}:
    \begin{itemize}
        \item If the loader returns a single \texttt{Trimesh}, we create a new mesh with the same vertices and faces (with \texttt{process=False} to avoid automatic repairs) and discard all materials.
        \item If the loader returns a \texttt{Scene}, we iterate over its geometries and extract each \texttt{Trimesh} in the same way.
        \item In both cases, we assign a uniform gray face color \texttt{[128, 128, 128, 255]} to decouple our geometry pipeline from the original textures. Later, we re-render per-part images using our own lighting and camera setup (Sec.~\ref{sec:appendix-data-aug}), so we do not rely on baked-in materials.
    \end{itemize}
    \item We collect all such geometry-only meshes for link $\ell$ into a list. If the list is empty, the link is effectively dropped.
    \item If there is exactly one mesh, we keep it as-is. If there are multiple, we merge them by explicit vertex-face concatenation:
    \begin{align}
        V_{\text{all}} &= 
        \begin{bmatrix}
            V_1 \\ \vdots \\ V_m
        \end{bmatrix}, \\
        F_{\text{all}} &=
        \begin{bmatrix}
            F_1 \\
            F_2 + |V_1| \\
            \vdots \\
            F_m + \sum_{i=1}^{m-1} |V_i|
        \end{bmatrix},
    \end{align}
    where $V_i$ and $F_i$ are the vertices and faces of the $i$-th sub-mesh and $|\cdot|$ denotes the number of vertices. We then create a single \texttt{Trimesh} from $(V_{\text{all}}, F_{\text{all}})$ with a uniform gray color.
\end{enumerate}

This procedure yields one rigid mesh per URDF link, with all subcomponents fused into the same local frame. We intentionally do not perform heavy processing at this stage (e.g., no automatic repairs or decimation) to preserve the original geometry as much as possible; watertight voxelization and cleanup are deferred to the next stage (Sec.~\ref{sec:watertight-preprocess}).

\section{Watertight Part Preprocessing}
\label{sec:watertight-preprocess}

Our network operates on per-part surface samples extracted from mesh geometry (Sec.~\ref{sec:network-appendix}). In practice, raw CAD / reconstruction data often contain small gaps, self-intersections, or open boundaries, which lead to unstable surface sampling and inconsistent volumes. Before training, we therefore convert every part mesh into a watertight surface via voxelization and marching cubes, combined with thin-part guards and aggressive post-cleaning. Unless otherwise stated, all experiments use a target voxel resolution of $R{=}200$ along the largest object axis.

\subsection{Per-part Voxelization and Pitch Selection}

Given a per-part mesh with axis-aligned bounding box extents
\[
    \mathbf{e} = (e_x, e_y, e_z) \in \mathbb{R}^3_{>0},
\]
we first choose a voxel pitch $\Delta$ that balances three goals: (1) roughly $R{=}200$ voxels along the largest axis, (2) at least $N_{\min}$ voxels across the thinnest axis to avoid losing doors and sheet-like structures, and (3) a soft memory budget for the voxel grid.

Concretely, we define two candidate pitches
\begin{equation}
    \Delta_{\text{res}} = \frac{\max(\mathbf{e})}{R},
    \qquad
    \Delta_{\text{thin}} = \frac{\min(\mathbf{e})}{N_{\min}},
\end{equation}
where $N_{\min}$ is a small integer (we use $N_{\min}{=}3$ in all experiments). The first term enforces the user-specified resolution along the largest axis; the second guarantees a minimum number of cells across the thinnest axis. We adopt the more conservative (higher-resolution) pitch
\begin{equation}
    \Delta = \min(\Delta_{\text{res}},\, \Delta_{\text{thin}}).
\end{equation}

Given $\Delta$, the approximate grid shape is
\begin{equation}
    \mathbf{n} = (n_x, n_y, n_z)
    =
    \left(
        \left\lceil \frac{e_x}{\Delta} \right\rceil,
        \left\lceil \frac{e_y}{\Delta} \right\rceil,
        \left\lceil \frac{e_z}{\Delta} \right\rceil
    \right),
\end{equation}
and we estimate memory usage as
\begin{equation}
    M_{\text{vox}} \approx n_x n_y n_z \;\text{bytes}.
\end{equation}
If $M_{\text{vox}}$ exceeds a rough cap (400\,MB in our implementation), we relax the pitch by a global scale factor so that $M_{\text{vox}}$ fits into the budget. This procedure yields a per-part voxel grid that is fine enough for thin structures but remains tractable even for large assets.

\subsection{Closed Occupancy and Marching Cubes}

With the chosen pitch $\Delta$, we voxelize each part mesh into an occupancy grid by calling the \texttt{voxelized} interface from \texttt{trimesh}. To enforce watertightness, we explicitly convert the occupancy grid into a solid by filling interior and small holes:
\begin{enumerate}
    \item Voxelize the original mesh at pitch $\Delta$.
    \item Apply a fill operation to propagate occupancy to the interior, producing a fully closed solid voxel grid.
\end{enumerate}
We then extract an isosurface using marching cubes on this filled grid. Because the occupancy is explicitly closed before marching cubes, the resulting surface is a watertight shell up to discretization artifacts.

\subsection{Axis-wise Rescaling Back to Original Extents}

The marching-cubes shell lives in the coordinate frame of the voxel grid, and its bounding box extents can deviate slightly from the original mesh due to discretization. To avoid systematic shrinkage of thin parts, we apply an anisotropic rescaling that exactly matches the original axis-aligned extents.

Let $\mathbf{e}^{\text{orig}}$ denote the original mesh extents and $\mathbf{e}^{\text{shell}}$ the extents of the marching-cubes shell. We compute a per-axis scale
\begin{equation}
\begin{aligned}
\mathbf{s} &= (s_x, s_y, s_z),\\
s_i &= \frac{e^{\text{orig}}_i}{\max\!\big(e^{\text{shell}}_i,\varepsilon\big)}, 
\quad i \in \{x,y,z\}.
\end{aligned}
\end{equation}
with a small $\varepsilon$ to avoid division by zero. We then:
\begin{enumerate}
    \item Translate the shell so that its bounding-box center is at the origin.
    \item Apply the diagonal scale matrix $\mathrm{diag}(\mathbf{s})$.
    \item Translate back so that the shell is centered at the original mesh center.
\end{enumerate}
This axis-wise rescale ensures that each reconstructed part exactly matches the original size along all three axes, preserving joint clearances and articulated contact patterns.

\subsection{Robust Mesh Cleanup and Early Exit for Watertight Parts}

After rescaling, we run a dedicated cleanup routine to remove numerical artifacts introduced by voxelization:
\begin{itemize}
    \item \textbf{Vertex welding.} We merge vertices within a small tolerance (we use a weld radius of $10^{-6}$ in world units) to eliminate near-degenerate triangles.
    \item \textbf{Degenerate and duplicate faces.} We repeatedly remove duplicate and zero-area faces, then drop unreferenced vertices and fix normals.
    \item \textbf{Largest connected component.} To discard floating fragments, we split the mesh into connected components and retain only the largest one (preferring components with more faces and, when available, larger volume).
    \item \textbf{Optional decimation.} For extremely dense outputs, we optionally apply quadratic decimation toward a target face count. For the experiments reported in this paper, we disable decimation (target faces set to zero) to avoid erasing small but semantically important details.
\end{itemize}

For parts that are already watertight and reasonably clean in the raw data, we take a conservative path: we skip voxelization and only apply the lightweight cleanup routine above. This preserves the original high-frequency geometry while still enforcing a consistent, watertight representation for problematic parts.

\section{Articulation-Aware Data Augmentation}
\label{sec:appendix-data-aug}

After constructing watertight per-part meshes and canonical surface samples (Sec.~\ref{sec:watertight-preprocess}), we augment the dataset by rendering each articulated object at additional joint configurations while keeping the underlying geometry and part decomposition fixed. Concretely, for every \texttt{mobility.urdf} with at least one revolute joint, we synthesize two extra variants:
\begin{itemize}
    \item a \emph{max} pose, where all revolute joints are placed at their upper joint limits; and
    \item a \emph{mid} pose, where all revolute joints are placed at half of their maximum angle.
\end{itemize}
Each variant reuses the same watertight mesh and part-level surface samples but provides a new RGB rendering at a different articulation state. This encourages the model to become robust to joint motion while still reconstructing a consistent canonical geometry for each object.

\subsection{URDF-Based Multi-Joint Forward Kinematics}

We rely on the original PartNet-Mobility URDF to recover kinematic structure. For each object we parse:
\begin{itemize}
    \item all links $\mathcal{L}$, each with one or more \texttt{<visual>} blocks that specify a mesh filename, an origin translation $\mathbf{t}^{\text{vis}} \in \mathbb{R}^3$, and an RPY rotation $\mathbf{r}^{\text{vis}} \in \mathbb{R}^3$;
    \item all joints $\mathcal{J}$, each connecting a parent link $p$ and child link $c$ with a joint origin $(\mathbf{t}^{\text{joint}}, \mathbf{r}^{\text{joint}})$, an axis $\mathbf{a} \in \mathbb{R}^3$, and a type (we only use \texttt{revolute} joints for augmentation).
\end{itemize}

For every link $\ell \in \mathcal{L}$, we choose a representative visual transform
\[
    T^{\text{vis}}_\ell \in \mathrm{SE}(3)
\]
by preferring GLB-based visuals when present and falling back to the first mesh visual otherwise. This defines the transform from the link frame to a canonical visual frame for that link.

We then perform visual-space forward kinematics over the kinematic tree. Root links are detected as links that never appear as a joint child. For each root link $r$, we initialize its visual transform in world coordinates as
\[
    T^{\text{world}}_{\text{vis}}(r) = T^{\text{vis}}_r,
    \qquad
    T^{\text{world}}_{\text{link}}(r) = I_4,
\]
so that its visual frame coincides with its world frame by construction.

For a joint $j \in \mathcal{J}$ with parent link $p$ and child link $c$, we form:
\begin{align}
    T^{\text{joint}}_j &= \mathrm{SE3}\big(R(\mathbf{r}^{\text{joint}}),\, \mathbf{t}^{\text{joint}}\big), \\
    T^{\text{rot}}_j(\theta_j) &= \mathrm{SE3}\big(R_{\text{axis}}(\mathbf{a}, \theta_j),\, \mathbf{0}\big),
\end{align}
where $R(\cdot)$ converts RPY to a rotation matrix and $R_{\text{axis}}(\mathbf{a}, \theta_j)$ is the Rodrigues rotation about axis $\mathbf{a}$ with angle $\theta_j$.

We propagate transforms along the kinematic tree in topological order using:
\begin{align}
    T^{\text{world}}_{\text{link}}(c)
    &= T^{\text{world}}_{\text{vis}}(p)\; T^{\text{joint}}_j\; T^{\text{rot}}_j(\theta_j), \\
    T^{\text{world}}_{\text{vis}}(c)
    &= T^{\text{world}}_{\text{link}}(c)\; T^{\text{vis}}_c.
\end{align}
This visual-space formulation matches the single-joint rendering script used during initial preprocessing and extends it to arbitrary joint depth without changing the relative placement of visual frames.

To align the multi-joint scene with the canonical coordinate system used in the main dataset, we pick a reference link $r^\star$ using a simple heuristic (prefer a fixed child of the base link; otherwise use the parent of the first revolute joint). Let $T^{\text{world}}_{\text{vis}}(r^\star)$ denote the computed visual transform and $T^{\text{vis}}_{r^\star}$ be its canonical visual transform. We apply a global similarity transform
\[
    S = T^{\text{vis}}_{r^\star} \big( T^{\text{world}}_{\text{vis}}(r^\star) \big)^{-1},
\]
and left-multiply all link and visual transforms by $S$. This guarantees that the reference link’s visual frame exactly matches the canonical preprocessing pipeline, while preserving all relative joint poses.

Finally, we instantiate a \texttt{trimesh.Scene} by loading every mesh referenced in the link visuals and applying the corresponding world transform
\[
    T^{\text{world}} = T^{\text{world}}_{\text{link}}(\ell)\; T^{\text{vis}}_{\ell,\text{component}}
\]
to each visual component. The result is a normalized, articulated scene in the same global coordinate system as the watertight mesh used for sampling.

\subsection{Sampling Joint Angles: Reference, Mid, and Max Poses}

For each revolute joint $j \in \mathcal{J}$, we read its upper limit $\theta^{\max}_j$ from the URDF \texttt{<limit>} tag when available, defaulting to $\theta^{\max}_j = \pi$ (180$^\circ$) if no limit is specified. We then define:
\[
    \theta^{\text{ref}}_j = 0, \qquad
    \theta^{\text{mid}}_j = \frac{1}{2}\,\theta^{\max}_j, \qquad
    \theta^{\text{max}}_j = \theta^{\max}_j.
\]

The original preprocessed dataset already contains the reference pose with all joints set to $\theta^{\text{ref}}_j$. Our augmentation script constructs two additional pose families:
\begin{description}
    \item[\textbf{Max pose} (\texttt{\_max}).] All revolute joints are set to $\theta^{\text{max}}_j$, simultaneously driving each joint to its mechanically allowed extreme. This exposes the model to highly opened drawers, doors, and other articulated components.
    \item[\textbf{Mid pose} (\texttt{\_mid}).] All revolute joints are set to $\theta^{\text{mid}}_j$, producing a configuration between the closed and fully open states. This captures typical everyday articulations without extreme self-occlusion.
\end{description}
Objects without any revolute joints are left unchanged and contribute only their reference pose.

\subsection{Consistent Normalization and Rendering Settings}

To ensure consistent scale and camera framing across all poses, we normalize each object once using the reference pose. Given the reference scene $\mathcal{S}_{\text{ref}}$ (all joints at $\theta^{\text{ref}}_j$), we compute its axis-aligned bounding box centroid $\mathbf{c}$ and extents $\mathbf{e}$. We then define:
\begin{align}
    \mathbf{t}_{\text{norm}} &= -\mathbf{c}, \\
    s_{\text{norm}} &= \frac{2}{\max(e_x, e_y, e_z)}.
\end{align}
We apply this normalization to all variants (reference, mid, max), so every object is centered at the origin and fits inside a unit cube regardless of articulation. Using a fixed normalization per object avoids small pose-induced scale changes and keeps camera parameters strictly comparable across augmentations.

For rendering, we reuse the same camera and lighting configuration as the main preprocessing pipeline:
\begin{itemize}
    \item camera placed on a sphere of fixed radius (4 units) around the origin;
    \item field of view of 40$^\circ$;
    \item high-resolution images at $2048 \times 2048$ pixels;
    \item an environment light setup with multiple evenly spaced directional lights (36 directions) and fixed intensity.
\end{itemize}
The augmentation script calls a shared \texttt{render\_single\_view} routine with these settings, so augmented images are visually indistinguishable from the original dataset except for joint angles.